\documentclass[10pt,twocolumn,letterpaper]{article}

\usepackage{cvpr}              %

\usepackage{colortbl}
\usepackage{bbding}

\definecolor{cvprblue}{rgb}{0.21,0.49,0.74}
\usepackage[breaklinks,colorlinks,allcolors=cvprblue]{hyperref}

\usepackage{svg}
\usepackage{multirow}

\def\paperID{4982} %
\def\confName{CVPR}
\def\confYear{2026}

\title{M$^4$-SAM: Multi-Modal Mixture-of-Experts with Memory-Augmented SAM for RGB-D Video Salient Object Detection}

\author{Jiyuan Liu$^1$, Jia Lin$^1$, Xiaofei Zhou$^{1,*}$, Runmin Cong$^2$, Deyang Liu$^3$, Zhi Liu$^4$\\
$^1$Hangzhou Dianzi University, China \quad
$^2$Shandong University, China \quad \\
$^3$Anqing Normal University, China \quad
$^4$Shanghai University, China\\
{\tt\small \{hankliu, lin\_j\}@hdu.edu.cn, zxforchid@outlook.com, rmcong@sdu.edu.cn,}\\
{\tt\small deyang.liu@hotmail.com, liuzhi@staff.shu.edu.cn}
}

\begin{document}
\maketitle
\renewcommand{\thefootnote}{\fnsymbol{footnote}}
\footnotetext[1]{Corresponding author.}
\begin{abstract}
\label{sec:abstract}
The Segment Anything Model 2 (SAM2) has emerged as a foundation model for universal segmentation.
Owing to its generalizable visual representations, SAM2 has been successfully applied to various downstream tasks.
However, extending SAM2 to the RGB-D video salient object detection (RGB-D VSOD) task encounters three challenges including
limited spatial modeling of linear LoRA, insufficient employment of SAM's multi-scale features, 
and dependence of initialization on explicit prompts.
To address the issues, we present \textbf{M}ulti-\textbf{M}odal \textbf{M}ixture-of-Experts with \textbf{M}emory-Augmented \textbf{SAM} (\textbf{M}$^4$-\textbf{SAM}), which equips SAM2 with modality-related PEFT, hierarchical feature fusion, and prompt-free memory initialization.
Firstly, we inject \textbf{Modality-Aware MoE-LoRA}, which employs convolutional experts to encode local spatial priors and introduces a modality dispatcher for efficient multi-modal fine-tuning, into SAM2's encoder.
Secondly, we deploy \textbf{Gated Multi-Level Feature Fusion}, which hierarchically aggregates multi-scale encoder features with an adaptive gating mechanism, to balance spatial details and semantic context.
Finally, to conduct zero-shot VSOD without manual prompts, we utilize a \textbf{Pseudo-Guided Initialization}, where a coarse mask is regarded as a pseudo prior and used to bootstrap the memory bank. 
Extensive experiments demonstrate that M$^4$-SAM achieves the state-of-the-art performance across all evaluation metrics on three public RGB-D VSOD datasets.

Code -- \href{https://github.com/HankLiu2020/M4-SAM}{https://github.com/HankLiu2020/M4-SAM}.
\end{abstract}
    
\section{Introduction}
\label{sec:intro}
Salient Object Detection (SOD) aims to identify and segment the most visually prominent objects that naturally attract human attention in a scene~\cite{borji2019salient}. It has been widely applied in various vision tasks, such as image/video segmentation~\cite{qin2019basnet,CRM}, robotics~\cite{bob2024quality} and defect detection~\cite{zxf2021dense,wb2023lfrnet}.
However, RGB-based SOD methods struggle in challenging scenarios, including complex backgrounds, poor illumination, and motion blur caused by fast movements. %

To tackle these limitations, researchers have explored incorporating additional modalities to provide complementary information. 
RGB-D SOD methods~\cite{PICRNet,hrtransnet,qu2017rgbd} leverage depth information to introduce robust geometric cues, while video SOD (VSOD) methods~\cite{zxf2023sti,wang2015consistent,ochs2013segmentation} introduce temporal information from video sequences to ensure consistent tracking and reduce motion ambiguity.
Building upon both modalities, RGB-D Video Salient Object Detection (RGB-D VSOD) methods~\cite{dvsod,rdvs,vidsod} leverage depth and temporal information simultaneously. %
However, most existing RGB-D VSOD methods are trained on limited-scale datasets, which restricts their generalization under real-world scenarios.

The recent emergence of Segment Anything Model (SAM)~\cite{sam1} has revolutionized the segmentation field with its powerful pre-trained representations and impressive zero-shot generalization capabilities.
Its successor SAM2~\cite{sam2} further extends its generalization to video segmentation via introducing built-in memory bank mechanisms. 
However, directly adapting SAM2 to RGB-D VSOD faces three critical challenges.
Firstly, the large model size and the limited scale of datasets make full-parameter fine-tuning infeasible.
Recent works~\cite{SAMed,sam2adapter,sam2unet,MDSAM} have explored incorporating Parameter-Efficient Fine-Tuning (PEFT)~\cite{PEFT} strategies, such as adapter~\cite{adapter} or LoRA~\cite{lora}.
However, vanilla LoRA employs only linear projections, lacking spatial priors and modality-specific design, thus limiting its ability to capture local structures and exploit cross-modal complementarity.
Secondly, most existing SAM-based methods only perform multi-scale fusion in the decoder through simple additive upsampling~\cite{MDSAM, shen2025mgd}, which restricts their ability to balance fine-grained spatial details and high-level semantic context.
Finally, memory-based methods~\cite{xmem,STM,stcn} typically rely on user-provided prompts (\emph{e.g.}, points, boxes, or masks) for first-frame prediction and memory initialization. Without such prompts, the memory bank faces an unreliable cold-start, hindering prompt-free RGB-D VSOD.

Motivated by the aforementioned issues, in this paper, we present a prompt-free framework, namely \textbf{M}ulti-\textbf{M}odal \textbf{M}ixture-of-Experts with \textbf{M}emory-Augmented \textbf{SAM} (\emph{\textbf{M}$^4$-\textbf{SAM}}), which adapts SAM2 for RGB-D video salient object detection. Our main contributions can be summarized as follows:

\begin{itemize}
\item We propose M$^4$-SAM to extend SAM2 to RGB-D VSOD, where the key components are modality-related PEFT, hierarchical feature fusion, and prompt-free memory initialization.

\item We design Modality-Aware MoE-LoRA, which elevates vanilla LoRA with convolutional experts and modality-specific routing, to conduct adaptive RGB-D feature fusion and efficient fine-tuning.

\item We develop a Pseudo-Guided Temporal Memory, which integrates a Gated Multi-Level Feature Fusion for hierarchical feature aggregation, 
and a Pseudo-Guided Initialization Strategy to eliminate manual prompt requirements using coarse masks as pseudo priors.

\item Extensive experiments on three RGB-D VSOD datasets demonstrate that M$^4$-SAM achieves state-of-the-art performance.
\end{itemize}

\section{Related Work}
\label{sec:related}

\subsection{SOD: From Images to RGB-D Videos}

\textbf{Salient Object Detection (SOD)} 
SOD~\cite{hou2017dss,han2017cnns,qin2019basnet,wang2019adaptive,pang2020multi,liu2021vst,zhuge2022icon,zxf2023transformer,STDNet} aims to detect and segment salient objects, which naturally attract human attention, from natural scenes~\cite{borji2019salient}.
Qin \emph{et al.}~\cite{qin2019basnet} propose BASNet with a densely supervised encoder-decoder network to achieve accurate boundary delineation.
Liu \emph{et al.}~\cite{liu2021vst} deploy a transformer architecture VST, which leverages multi-level token fusion and patch-task-attention.
However, RGB-based SOD encounters inherent limitations in several challenging scenarios, such as complex backgrounds, poor illumination, and motion blur caused by fast movements~\cite{zhou2021rgbdsurvey,fan2019shifting,tu2022rgbt}.

\textbf{RGB-D SOD} 
RGB-D SOD methods incorporate depth information as a complementary modality, providing geometric cues that are robust to appearance variations. 
Existing RGB-D methods can typically be categorized into three types~\cite{zhou2021rgbdsurvey}, namely early fusion, late fusion, and multi-scale fusion.
The early fusion methods~\cite{peng2014rgbd,zhao2020single,qu2017rgbd} concatenate cross-modal features at input or shallow layers, preserving low-level details but potentially losing modality-specific features.
The late fusion methods~\cite{han2017cnns,wang2019adaptive} process modalities independently until the decision-level prediction, extracting modality-specific features but missing cross-modal interactions.
Unlike these, the multi-scale fusion strategies~\cite{rdvs,PICRNet,hrtransnet} perform cross-modal interaction at multiple network layers, effectively balancing low-level details and high-level semantic information.
For instance, Cong \emph{et al.}~\cite{PICRNet} propose PICRNet, which employs a cross-modality point-aware interaction module to explore RGB-depth feature relationships under saliency guidance. Tang \emph{et al.}~\cite{hrtransnet} propose a two-stream HRFormer-based method HRTransNet, which integrates depth features as auxiliary features into the corresponding levels of the RGB branch.

\textbf{RGB-D VSOD} 
RGB-D VSOD further extends RGB-D SOD with temporal information, enabling consistent object segmentation across frames even under occlusions and motion blur.
For example, Li \emph{et al.}~\cite{dvsod} employ CRM~\cite{CRM} for layer-wise cross-modal fusion and STM~\cite{STM} for temporal aggregation, where temporal modeling is achieved through simple concatenation without task-specific architectural refinement.
Lin \emph{et al.}~\cite{vidsod} utilize optical flow for temporal modeling,
treating RGB, depth, and flow inputs equally through cross-level interactions.
Cho \emph{et al.}~\cite{TransFlow} leverage pre-trained video diffusion models to synthesize semantically-aware optical flows for temporal modeling.
Mou \emph{et al.}~\cite{rdvs} regard RGB as the primary modality, and propose a Universal Interaction Module (UIM) to concatenate high-level multi-modal features and generate a coarse map as a prior for refined feature fusion. %
However, these optical flow-based methods fail to capture long-term dependencies across multiple frames.
Meanwhile, they are typically trained on domain-specific datasets with limited scale, constraining generalization and requiring careful task-specific architectural design.

\subsection{SAM-based Segmentation Approaches}
Benefiting from massive-scale training datasets, Segment Anything Model (SAM)~\cite{sam1} and its successor SAM2~\cite{sam2} bring powerful pre-trained visual representations across diverse visual domains. %
However, their large model size, combined with the limited-scale task-specific downstream datasets, makes full-parameter fine-tuning infeasible.

Inspired by LLM fine-tuning, researchers have applied PEFT methods~\cite{PEFT} to SAM. 
For instance, Chen \emph{et al.}~\cite{sam2adapter} incorporate adapters~\cite{adapter} into SAM2's transformer blocks, enhancing its performance across multiple downstream tasks. 
Zhang \emph{et al.}~\cite{SAMed} apply LoRA~\cite{lora} to SAM's encoder, injecting low-rank matrices into the query and value projections of self-attention layers to adapt SAM for medical image segmentation. 
However, LoRA relies solely on linear projections, %
limiting its ability to model local spatial priors that are crucial for dense prediction tasks~\cite{convlora}.
In addition, the absence of modality-specific design forces multi-modal inputs to be processed through separate encoders, resulting in redundant computation and considerable memory overhead during the training stage.
To address this limitation, Zhong \emph{et al.}~\cite{convlora} propose Conv-LoRA, which incorporates MoE~\cite{moe2017} and convolutional layers into LoRA modules. %
In addition, Li \emph{et al.}~\cite{KANSAM} propose KAN-adapter for RGB-T SOD, realizing cross-modal fusion via element-wise addition and Kolmogorov Arnold Network layers within adapters~\cite{KAN}.

Recent works~\cite{MDSAM, shen2025mgd} have observed that SAM relies solely on the deepest-level features for prediction, which neglects the rich multi-level representations in the encoder.
To address this, Gao \emph{et al.}~\cite{MDSAM} propose a Multi-Level Fusion Module (MLFM) that aggregates encoder features from multiple layers using learnable weight distributors.
Shen \emph{et al.}~\cite{shen2025mgd} propose a Hierarchical Multi-view Interaction Module (HMIM), which progressively upsamples high-level semantic features, and fuses them with low-level spatial details to integrate global and local representations.

To extend SAM into video-based tasks, 
Deng \emph{et al.}~\cite{memsam} combine SAM with XMem~\cite{xmem}, utilizing first-frame prompts (\emph{e.g.}, points, boxes, or masks) for initial predictions, while the memory encoder fuses predicted masks with image features and stores them in a memory bank, enabling ultrasound video segmentation.
SAM2~\cite{sam2} further integrates a streaming memory, which propagates prompt embeddings and encoded features across the video sequence, achieving universal video segmentation. %
However, the reliance on user-provided prompts for first-frame prediction and memory initialization makes these methods unsuitable for the prompt-free RGB-D VSOD setting.

\section{Method}
\label{sec:method}

\begin{figure*}[t]
  \centering
  \includegraphics[width=0.88\textwidth]{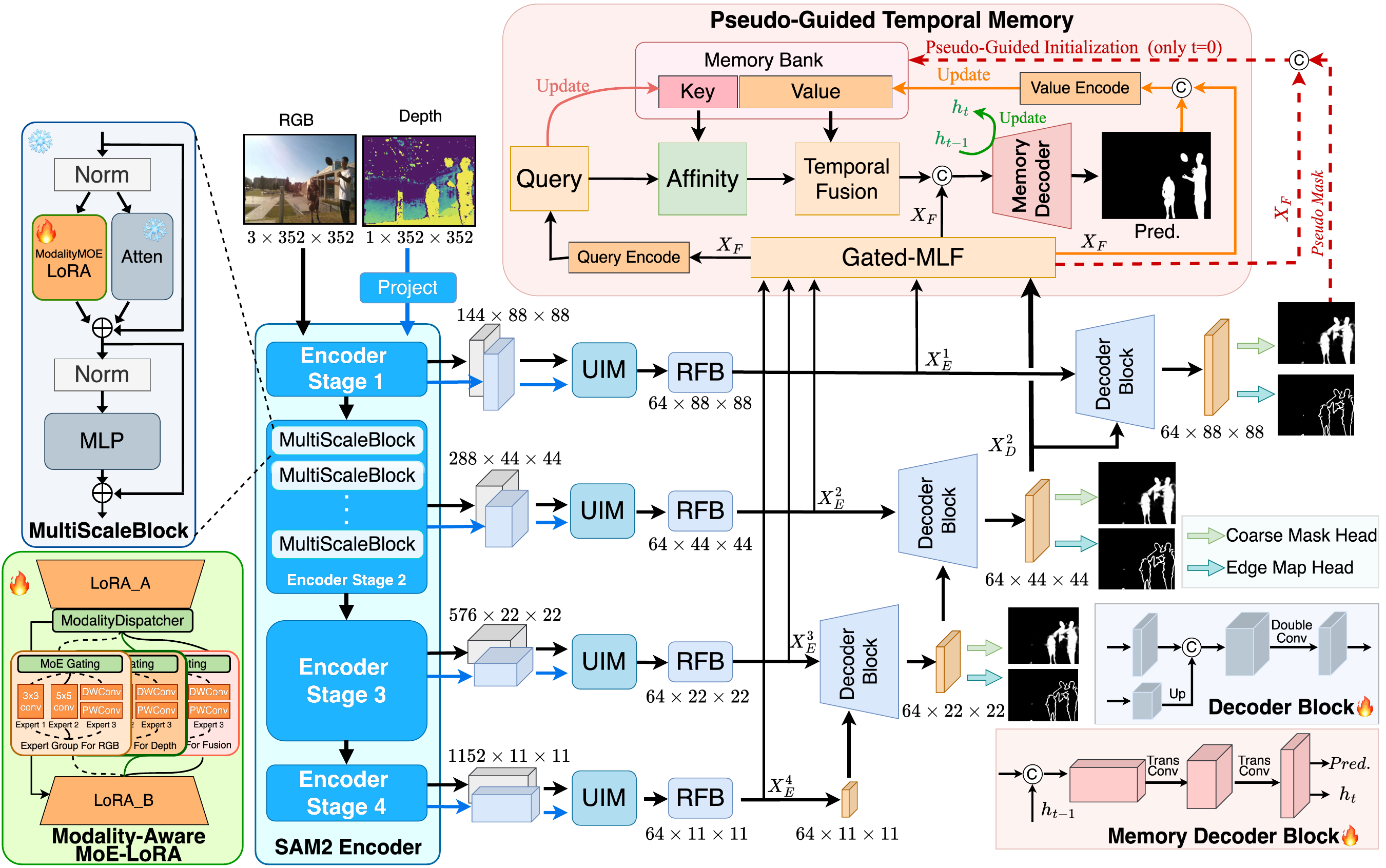}
  \caption{The overall architecture of our proposed Multi-Modal Mixture-of-Experts with Memory-Augmented SAM (M$^4$-SAM).}%
  \label{fig:overview}
  \vspace{-0.3cm}  
\end{figure*}

\subsection{Preliminary}

LoRA~\cite{lora} is a parameter-efficient fine-tuning (PEFT) technique, which is inserted into the query and value projection matrices within the self-attention layers.
It freezes the original projection matrices $W_0 \in \mathbb{R}^{d \times k}$, and injects a pair of trainable low-rank matrices $A \in \mathbb{R}^{r \times k}$ and $B \in \mathbb{R}^{d \times r}$.
The forward pass can be formulated as follows:
\begin{equation}
  h = W_0x + \Delta Wx = W_0x + BAx,
\end{equation}
where the rank constraint (\emph{i.e.}, $r \ll \min(d, k)$) significantly reduces computational cost and memory usage.
This low-rank adaptation allows efficient task-specific fine-tuning while preserving the strong generalization ability of the frozen foundation model.

\subsection{Overall Architecture}

As shown in Fig.~\ref{fig:overview}, our M$^4$-SAM reformulates SAM2 into a U-shaped architecture, %
enabling prompt-free RGB-D VSOD through hierarchical feature fusion while preserving its strong generalization ability.
Specifically, the RGB and depth inputs are processed by a shared Hiera encoder~\cite{hiera} augmented with the Modality-Aware MoE-LoRA, which employs modality-specific expert group activation to extract four-level features for each modality efficiently.
These features are then fused through the Universal Interaction Module (UIM)~\cite{rdvs} and the Receptive Field Block (RFB)~\cite{rfb}, combining complementary cross-modal information and producing unified multi-modal encoder features $\{X_E^i\}_{i=1}^4$.
After that, the encoder features are hierarchically decoded via a series of Decoder Blocks, yielding the decoder features $\{X_D^i\}_{i=1}^4$.
By integrating the middle-level decoder feature $X_D^2$ and the skip-connected encoder features $\{X_E^i\}_{i=1}^4$, the Pseudo-Guided Temporal Memory interacts the features of the current frame with the temporal information stored in the memory bank, and we can obtain the final high-quality prediction.
Finally, the memory bank is reversely updated using the latest prediction and corresponding features, to ensure temporal consistency and adapt to scene variations over time.

\subsection{Encoder with Modality-Aware MoE-LoRA}

To fully leverage the powerful capabilities of SAM's encoder, we introduce the Modality-Aware MoE-LoRA, which introduces mixture-of-experts (MoE) into the LoRA framework, enabling efficient adaptation and complementary feature extraction across RGB and depth modalities.

Unlike conventional LoRA, whose low-rank matrices are injected into attention projections via linear operations, we reformulate each LoRA branch into groups of convolutional experts.
Specifically, we design three types of convolutional experts including two standard convolution experts with kernel sizes of $3 \times 3$ and $5 \times 5$, and one composed of depthwise and pointwise convolutions for efficient inference.
These convolutional experts introduce spatial locality priors that better align with visual structure, enabling efficient fine-tuning for image-based tasks.
Moreover, the combination of different kernel sizes enables the model to capture diverse spatial patterns across varying receptive fields.
After that, the outputs of these experts are adaptively aggregated through a lightweight MoE Gating module, which dynamically selects and combines the top-$K$ relevant experts based on the input representation.

Considering the inherent discrepancy between RGB and depth modalities, we further extend the MoE formulation and introduce three different groups, including RGB, depth, and fusion.
The modality-specific groups (\emph{i.e.}, RGB and depth) highlight the intra-modal representations, while the fusion group enables multi-level cross-modal interaction.
To achieve adaptive activation of the expert groups, we design a Modality Dispatcher $\mathbf{D}(\cdot)$, which dynamically routes features to corresponding expert groups based on the input modality.
Specifically, the dispatcher activates RGB and fusion groups for the RGB stream, and depth and fusion groups for the depth stream, where the fusion group parameters are shared across both streams.
Unlike standard PEFT methods (\emph{e.g.}, Adapter~\cite{adapter} and LoRA~\cite{lora}) that lack dynamic modeling capabilities and thus require separate encoder branches for additional modalities, the Modality Dispatcher enables RGB and depth inputs to be processed within a unified encoder branch, 
significantly reducing memory overhead while maintaining modality-specific representations.

The forward pass of our Modality-Aware MoE-LoRA can be formulated as follows:
\begin{equation}
  h = W_0x + \Delta W x = W_0x + BAx + B\mathbf{D}(Ax),
\end{equation}
where $\mathbf{D}(\cdot)$ is the Modality Dispatcher, which routes the low-rank features $Ax$ to either RGB+fusion or depth+fusion expert groups based on the current input modality.

Overall, our encoder follows a hierarchical structure that allows multi-scale feature extraction.
Specifically, the single encoder with Modality-Aware MoE-LoRA processes both RGB and depth inputs through modality-specific expert activation, generating the multi-level RGB features $\{X^{i}_{RGB}\}_{i=1}^4$ and depth features $\{X^{i}_{depth}\}_{i=1}^4$.
After that, the multi-level RGB and depth features are fused via the Universal Interaction Module (UIM)~\cite{rdvs} and the Receptive Field Block (RFB)~\cite{rfb}, yielding the unified encoder features $\{X_E^i\}_{i=1}^4$.

\subsection{Hierarchical Decoder}
Following the multi-scale encoder features $\{X_E^i\}_{i=1}^4$, we employ a hierarchical decoder to progressively reconstruct spatial details.
Each decoder block performs upsampling and skip fusion with the corresponding encoder feature, producing a series of decoder representations $\{X_D^i\}_{i=1}^4$.
To optimize the reconstruction and promote the boundary accuracy,
we feed the decoder features into two lightweight heads, where one generates coarse segmentation masks $P_{c}^i$ and the other produces edge maps $P_{e}^i$. 
This auxiliary supervision design enables multi-scale refinement and preserves spatial precision while maintaining semantic consistency.

\subsection{Pseudo-Guided Temporal Memory}

To enhance temporal stability and generate consistent predictions across frames, we propose a Pseudo-Guided Temporal Memory, which consists of two key components, namely a Gated Multi-Level Feature Fusion module for multi-scale feature aggregation and a Memory Bank for temporal modeling.

\begin{figure}[t]
  \centering
  \includegraphics[width=0.5\textwidth]{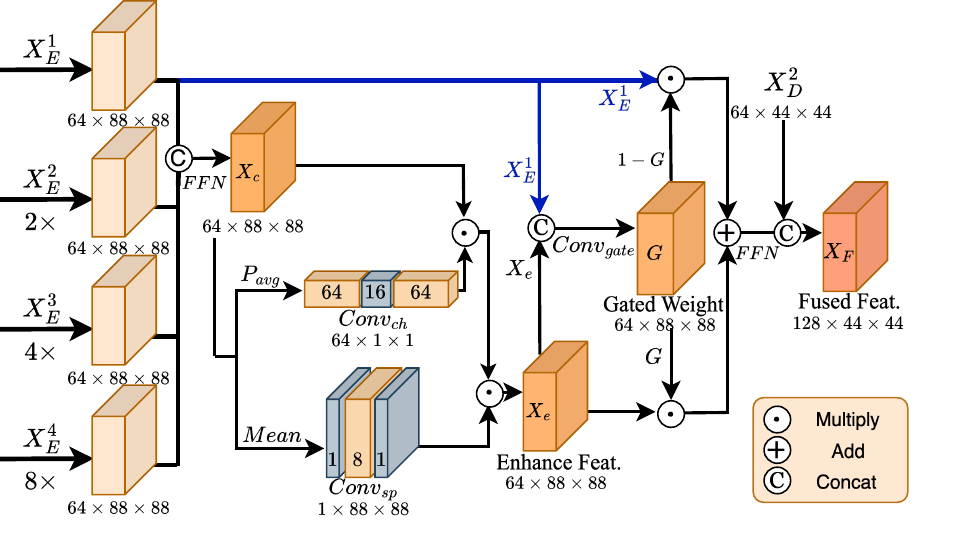}
  \caption{The Gated Multi-Level Feature Fusion module.}
  \label{fig:mlf}
  \vspace{-0.3cm}  
\end{figure}

As shown in Fig.~\ref{fig:mlf}, given the multi-scale encoder features $\{X_E^i\}_{i=1}^4$, we concatenate and compress them through an FFN into a unified context representation $X_{c}$.
To enhance spatial-channel interactions, $X_{c}$ is refined by dual attention operations, namely
\begin{equation}
  X_e = Conv_{sp}(Mean(X_{c})) \cdot [Conv_{ch}(P_{avg}(X_{c})) \cdot X_c],
\end{equation}
where $Conv_{sp}$ and $Conv_{ch}$ denote spatial and channel convolutions respectively, $Mean$ denotes channel-wise averaging ($C{\times}H{\times}W \to 1{\times}H{\times}W$), and $P_{avg}$ denotes global average pooling ($C{\times}H{\times}W \to C{\times}1{\times}1$).
Then, a gated weight $G$ adaptively balances the shallow and enhanced features, generating the fused encoder representation $\tilde{X}_E$:
\begin{equation}
  \left \{
    \begin{array}{l}
      G = Conv_{gate}(Concat(X_E^1, X_e)) \\  %
      \tilde{X}_E = FFN(G \cdot X_e + (1 - G) \cdot X_E^1)
    \end{array},
  \right.
\end{equation}
where $Conv_{gate}$ is a lightweight convolutional layer along the channel dimension that produces element-wise gating weights.
By integrating the fused encoder representation and decoder feature, Gated-MLF generates the fused representation $X_F$, which can be written as follows:
\begin{equation}
  X_F = Concat(\tilde{X}_E, X_D^2).
\end{equation}
Note that, inspired by recent observations~\cite{shao2024explore} that the final layer of a foundation model tends to suppress local spatial information, making it less suitable for dense prediction tasks, we employ the middle-level decoder feature $X_D^2$ rather than the final one for fusion.

After that, to incorporate temporal cues, we perform the cross-attention between $X_F$ and the memory bank, namely
\begin{equation}
  \tilde{X}_F = softmax(\frac{Q \cdot K_m^T}{\sqrt{d}}) \cdot V_m,
\end{equation}
where $Q$ is the query projected from current frame features $X_F$. $K_m$ and $V_m$ are the keys and values projected from previous frames, which are stored in the memory bank. In this way, we can obtain the temporally aggregated feature $\tilde{X}_F$.
Finally, the memory decoder processes the spatial feature $X_F$, temporal context $\tilde{X}_F$, and previous hidden state $h_{t-1}$, yielding the prediction mask $P$.

After obtaining the prediction, we update the memory bank with the current prediction and frame features.
Given the query embedding $Q_t$, the fused features $X_{F, t}$ and corresponding prediction $P_t$ at frame $t$, we generate the new key element $k_{m, t}$ and value element $v_{m, t}$, which can be written as follows:
\begin{equation}
  \left \{
    \begin{array}{l}
      k_{m, t} = Q_t \\
      v_{m, t} = \text{ValueEncoder}(X_{F, t}, P_t)
    \end{array},
  \right.
\end{equation}
where $\text{ValueEncoder}$ is a linear projection operation that filters and encodes salient features under the guidance of the prediction mask. 
By incorporating the $k_{m, t}$ and $v_{m, t}$, memory bank stores the latest $T$-frames temporal keys and values elements, namely $K_{m, t+1} = \{k_{m, t}, k_{m, t-1}, \ldots, k_{m, t-T+1}\}$ and $V_{m, t+1} = \{v_{m, t}, v_{m, t-1}, \ldots, v_{m, t-T+1}\}$, enabling temporal consistency across the video sequence.

Since there is no available temporal information for the first frame, we propose a Pseudo-Guided Initialization strategy, which utilizes the coarse mask as a pseudo mask to initialize the memory bank. This is considerably different from SAM2~\cite{sam2} and XMem~\cite{xmem}, which either employ the user-provided prompts or utilize ground truth at the first frame. 
Specifically, the pseudo mask is passed through two linear projections, namely
\begin{equation}
  \left \{
    \begin{array}{l}
      \tilde{k}_{m, 0} = \text{Linear}_k(X_{F, 0}) \\
      \tilde{v}_{m, 0} = \text{Linear}_v(X_{F, 0} \cdot P_{c, 0}^1)
    \end{array},
  \right.
\end{equation}
where $\tilde{k}_{m, 0}$ and $\tilde{v}_{m, 0}$ denote the initial memory key and memory value, $\text{Linear}_k$ and $\text{Linear}_v$ are the key and value projection layers, and $P_{c, 0}^1$ denotes the coarse mask generated from $X_D^1$ at the first frame.
Note that, the value projection shares parameters with the value encoder, which is used to maintain a unified feature space between initialization and subsequent temporal modeling.

This pseudo-guided initialization removes the need for manual-provided prompts, and extracts the stable temporal representations at the first frame.
Moreover, since the memory bank relies on attention mechanism, noisy initial keys naturally yield low affinity scores, suppressing erroneous weights and preventing error propagation.

\begin{table*}
    \centering
    \caption{Comparison of state-of-the-art methods on RGB-D video salient object detection benchmarks. The best three results in each column are marked in \textcolor{red}{red}, \textcolor{green}{green}, and \textcolor{blue}{blue}, respectively.}
    \label{tab:comparison}
    \resizebox{1\textwidth}{!}{%
      \begin{tabular}{l|c|cccc|cccc|cccc}
      \hline
      \multirow{2}{*}{Methods} &
      \multirow{2}{*}{Reference} &
      \multicolumn{4}{c|}{DViSal} &
      \multicolumn{4}{c|}{RDVS} &
      \multicolumn{4}{c}{ViDSOD-100} \\
      \cline{3-14}
      & & %
      $E_\xi \uparrow$ & $S_\alpha \uparrow$ & $F_\beta \uparrow$ & $M \downarrow$ &
      $E_\xi \uparrow$ & $S_\alpha \uparrow$ & $F_\beta \uparrow$ & $M \downarrow$ &
      $E_\xi \uparrow$ & $S_\alpha \uparrow$ & $F_\beta \uparrow$ & $M \downarrow$ \\
      \hline
      MFENet & ICASSP'25 &%
      - & 0.760 & 0.717 & 0.080 &
      - & 0.794 & 0.700 & 0.049 &
      - & 0.831 & 0.763 & 0.040 \\
      PICRNet & MM'23 &%
      0.715 & 0.670 & 0.568 & 0.147 &
      0.743 & 0.728 & 0.535 & 0.074 &
      0.873 & 0.830 & 0.738 & 0.038 \\
      STDNet & TPAMI'25 &%
      0.727 & 0.646 & 0.505 & 0.156 &
      0.673 & 0.595 & 0.311 & 0.089 &
      0.773 & 0.698 & 0.562 & 0.078 \\
      HRTransNet & TCSVT'22 &%
      0.727 & 0.628 & 0.497 & 0.161 &
      0.725 & 0.671 & 0.445 & 0.076 &
      0.745 & 0.686 & 0.531 & 0.099 \\
      ATFNet & IJCV'24 &%
      0.795 & 0.724 & 0.622 & 0.111 &
      0.732 & 0.713 & 0.491 & 0.074 &
      0.901 & 0.875 & 0.813 & 0.027 \\
      DPA & CVPR'24 &%
      0.796 & 0.724 & 0.635 & 0.102 &
      0.675 & 0.666 & 0.445 & 0.096 &
      0.848 & 0.817 & 0.715 & 0.051 \\
      DVSOD & NeurIPS'23 &%
      0.807 & 0.729 & 0.610 & 0.113 &
      0.748 & 0.587 & 0.452 & 0.070 &
      0.783 & 0.702 & 0.568 & 0.083 \\
      DCTNet+ & TIP'24 &%
      0.828 & 0.767 & 0.689 & 0.095 &
      \textcolor{green}{0.909} & \textcolor{green}{0.876} & \textcolor{green}{0.794} & \textcolor{blue}{0.029} &
      0.901 & 0.876 & 0.809 & 0.030 \\
      LSTA & PR'24 &%
      0.848 & 0.700 & 0.640 & 0.082 &
      0.746 & 0.650 & 0.484 & 0.069 &
      0.757 & 0.671 & 0.565 & 0.086 \\
      MDSAM & MM'24 &%
      0.856 & 0.796 & 0.715 & 0.071 &
      0.813 & 0.791 & 0.647 & 0.056 &
      \textcolor{blue}{0.909} & 0.877 & 0.815 & \textcolor{blue}{0.026} \\
      SAM2-UNet & Vis. Intell. &%
      0.856 & 0.778 & \textcolor{blue}{0.747} & \textcolor{blue}{0.064} &
      \textcolor{blue}{0.888} & 0.843 & 0.765 & 0.035 &
      0.907 & \textcolor{blue}{0.891} & \textcolor{blue}{0.829} & \textcolor{green}{0.025} \\
      TransFlow & ICCVW'25 &%
      \textcolor{blue}{0.866} & \textcolor{blue}{0.798} & 0.721 & 0.070 &
      0.876 & 0.840 & 0.707 & 0.035 &
      0.898 & 0.874 & 0.799 & 0.031 \\
      KAN-SAM & ICME'25 &%
      \textcolor{green}{0.885} & \textcolor{green}{0.835} & \textcolor{green}{0.783} & \textcolor{green}{0.052} &
      \textcolor{blue}{0.888} & \textcolor{blue}{0.854} & \textcolor{blue}{0.791} & \textcolor{green}{0.028} &
      \textcolor{green}{0.912} & \textcolor{green}{0.892} & \textcolor{green}{0.846} & \textcolor{green}{0.025} \\
      \rowcolor{red!20}
      M$^4$-SAM & Ours &%
      \textcolor{red}{0.925} & \textcolor{red}{0.850} & \textcolor{red}{0.828} & \textcolor{red}{0.039} &
      \textcolor{red}{0.927} & \textcolor{red}{0.878} & \textcolor{red}{0.802} & \textcolor{red}{0.027} &
      \textcolor{red}{0.936} & \textcolor{red}{0.912} & \textcolor{red}{0.868} & \textcolor{red}{0.016} \\
      \hline
      \end{tabular}%
    }
    \vspace{-0.3cm}  
  \end{table*}

\subsection{Loss Function}

We train our model with the following loss function, namely
\begin{equation}
\mathcal{L}_{total} = \mathcal{L}_{pred} + \mathcal{L}_{aux} + \mathcal{L}_{moe},
\end{equation}
where $\mathcal{L}_{pred}$, $\mathcal{L}_{aux}$ and $\mathcal{L}_{moe}$ denote final prediction loss, multi-level auxiliary loss and MoE regularization loss, respectively.

\noindent{\bf Final Prediction Loss.}
We introduce the structure loss~\cite{F3Net} to reduce the discrepancies between the final prediction and the ground truth (GT).

\noindent{\bf Multi-level Auxiliary Loss.}
We apply intermediate supervision to decoder features, which is formulated as:
\begin{equation}
  \mathcal{L}_{aux} = \sum_{i=1}^{3}(\mathcal{L}_{pred_c}^i + \mathcal{L}_{edge}^i),
\end{equation}
where $\mathcal{L}_{pred_c}^i$ and $\mathcal{L}_{edge}^i$ respectively denote the losses for coarse masks $P_{c}^i$ (supervised by GT) and edge maps $P_{e}^i$ (supervised by Sobel-filtered pseudo edges) at the $i$-th level.

\noindent{\bf MoE Regularization.}
Following~\cite{moe2017}, we adopt the load-balancing loss $\mathcal{L}_{moe} = \lambda [(\sigma(\mathbf{I})/\mu(\mathbf{I}))^2 + (\sigma(\mathbf{L})/\mu(\mathbf{L}))^2]$ with $\lambda{=}10^{-2}$ to ensure uniform expert utilization and prevent expert collapse, where $\mathbf{I}$ and $\mathbf{L}$ denote the importance and load of each expert.

\section{Experiments}
\label{sec:experiments}

\subsection{Dataset and benchmarks}

To conduct a comprehensive validation of our method, we train and test our model on three RGB-D video salient object detection benchmarks.

\noindent{\bf DViSal}~\cite{dvsod} consists of 237 RGB-D videos with 7117 frames. It includes both indoor and outdoor scenes with complex illumination conditions.

\noindent{\bf RDVS}~\cite{rdvs} contains 57 RGB-D videos with 4087 frames captured in diverse real-world scenarios, containing complex backgrounds and various object movements. 

\noindent{\bf ViDSOD-100}~\cite{vidsod} comprises 100 RGB-D videos with 9362 frames in total. This dataset presents challenges including scale variations and fast motion.

\subsection{Experimental Setup}

\noindent{\bf Evaluation Metrics.}
We adopt four widely-used metrics for comprehensive evaluation:
{E-measure} ($E_\xi$)~\cite{emeasure}, {S-measure} ($S_\alpha$)~\cite{smeasure}, {F-measure} ($F_\beta$)~\cite{fmeasure}, and {Mean Absolute Error} (MAE, $M$).
For $E_\xi$, $S_\alpha$ and $F_\beta$, higher values indicate better performance, while a lower $M$ denotes higher accuracy.

\noindent{\bf Implementation Details.}
We adopt the large-scale Hiera encoder from SAM2.1 (SAM-L)~\cite{sam2}, pre-trained on SA-1B, as the backbone. Within the frozen encoder, only the injected Modality-Aware MoE-LoRA modules are trainable for efficient fine-tuning. Each MoE-LoRA block uses low-rank matrices with rank $r=4$, incorporating 3 convolutional experts per group and activating the top-$2$ most relevant experts via the MoE Gating.
Training is performed using AdamW with separate learning rates of $1{\times}10^{-4}$ for MoE-LoRA and $1{\times}10^{-3}$ for other parameters, and a weight decay of $5{\times}10^{-4}$. We train the model for 50 epochs on two NVIDIA RTX 4090 GPUs (48GB) with a batch size of 4 per GPU.
Each training clip contains $T{=}4$ consecutive RGB-D frames resized to $352{\times}352$ (depth maps normalized to $[0, 1]$ and replicated to 3 channels), where the memory bank progressively accumulates temporal context across $T{-}1$ frames within the clip. Owing to the lightweight design of Modality-Aware MoE-LoRA, the entire training process can be completed within 5 hours.

\begin{figure*}[t]
  \centering
  \includegraphics[width=0.90\textwidth]{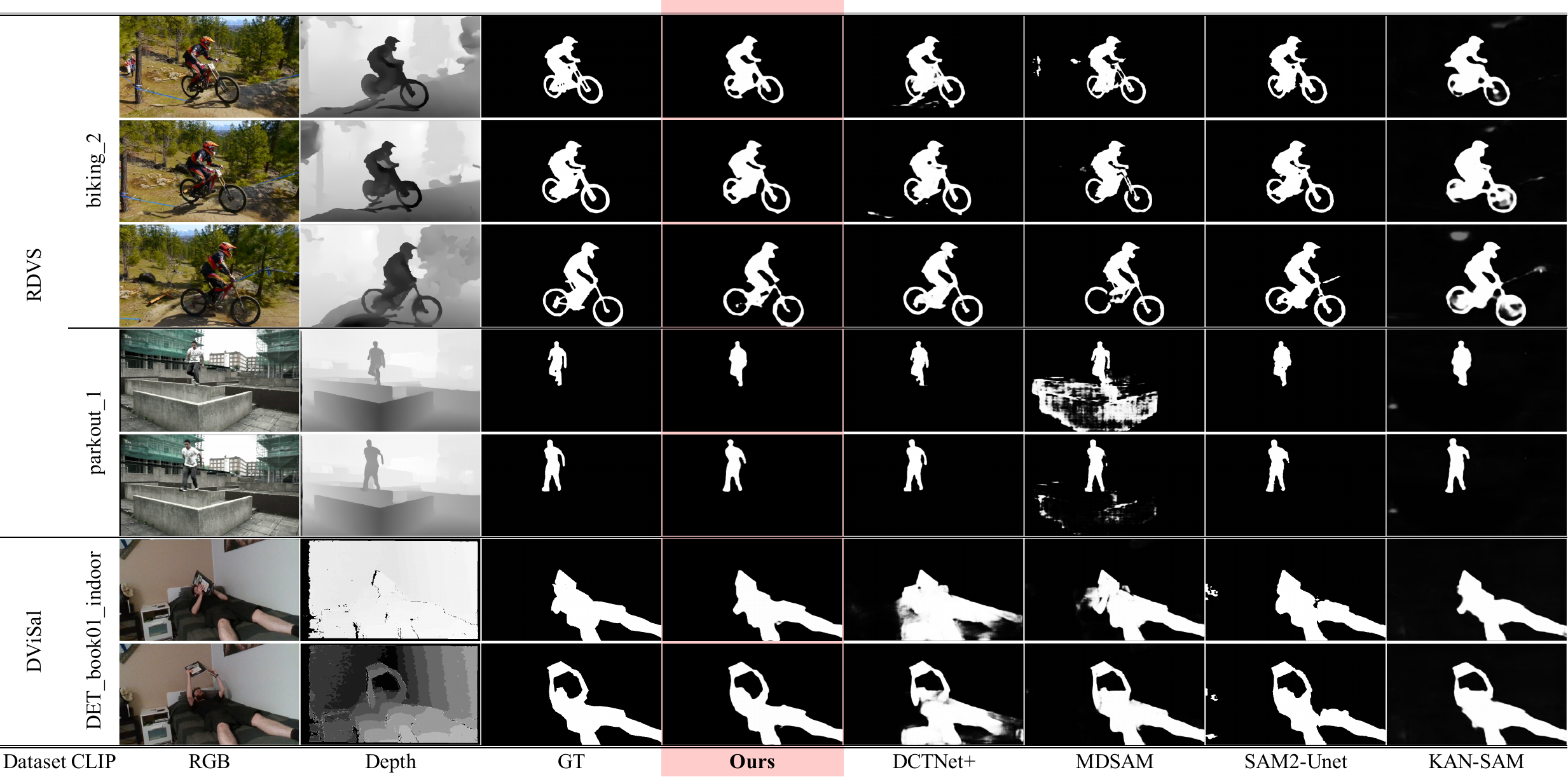}
  \caption{Qualitative comparison with representative state-of-the-art models.}
  \label{fig:qualitative_comparison}
  \vspace{-0.3cm}  
\end{figure*}

\subsection{Comparison with State-of-the-art}

As shown in Table~\ref{tab:comparison}, we quantitatively compare M$^4$-SAM against 13 state-of-the-art methods including %
MFENet~\cite{MFENet}, PICRNet~\cite{PICRNet}, STDNet~\cite{STDNet}, HRTransNet~\cite{hrtransnet}, ATFNet~\cite{vidsod}, DPA~\cite{DPA}, DVSOD~\cite{dvsod}, DCTNet+~\cite{rdvs}, LSTA~\cite{LSTA}, MDSAM~\cite{MDSAM}, SAM2-UNet~\cite{sam2unet}, TransFlow~\cite{TransFlow}, and KAN-SAM~\cite{KANSAM}. Our method achieves the best performance across all evaluation metrics on all datasets.
Specifically, on the DViSal dataset, our method reaches 0.925 and 0.828 in terms of E-measure and F-measure, achieving 4.5\% and 5.7\% improvements over the second-best method KAN-SAM, respectively.
On the RDVS dataset, M$^4$-SAM obtains 0.927 in terms of E-measure, surpassing the second-best method DCTNet+ by 2.0\%.
On the ViDSOD-100 dataset, we achieve 0.936 and 0.016 in terms of E-measure and MAE, outperforming the second-best method KAN-SAM by 2.6\% and 0.009, respectively.
Besides, to demonstrate that the performance gain of our method is not solely due to the SAM2 backbone, we make comparisons with various SAM-based baselines (\emph{i.e.}, MDSAM, SAM2-UNet, and KAN-SAM).
Compared to SAM-based methods, M$^4$-SAM achieves average E-measure improvements of 6.9\%, 7.6\%, and 2.9\% respectively across all three datasets.

\begin{table}
  \centering
  \caption{Ablation studies of Depth Input.}
  \label{tab:ablation_depth_input}
  \resizebox{0.85\columnwidth}{!}{%
  \begin{tabular}{l|cccc}
    \hline
    Depth Input & $E_\xi \uparrow$ & $S_\alpha \uparrow$ & $F_\beta \uparrow$ & $M \downarrow$ \\
    \hline
    Pseudo (Copy) & 0.858 & 0.740 & 0.623 & 0.047 \\
    Pseudo (Black) & 0.895 & 0.857 & 0.768 & 0.029 \\
    \rowcolor{red!20}
    \textbf{Actual (Ours)} & \textbf{0.927} & \textbf{0.878} & \textbf{0.802} & \textbf{0.027} \\
    \hline
  \end{tabular}
  }
  \vspace{-0cm}  
\end{table}

The qualitative comparison is presented in Fig.~\ref{fig:qualitative_comparison}, which compares our M$^4$-SAM with DCTNet+ and other SAM-based methods on representative video sequences. %
Here, we present three challenging scenarios, including a fast motion scene (biking\_2, rows 1-3), a complex background scene (parkout\_1, rows 4-5), and a low-light indoor scene (DET\_book01\_indoor, rows 6-7).
Specifically, DCTNet+, MDSAM and KAN-SAM suffer from inaccurate boundaries in fast motion scene due to their limited temporal modeling capability.
In complex background environments, MDSAM struggles to distinguish salient objects from cluttered backgrounds, as it relies solely on RGB cues without depth guidance.
In low-light indoor scene, DCTNet+ and SAM2-UNet fail to assess the reliability of depth information, leading to inaccurate predictions when the depth maps are noisy.
In contrast, our M$^4$-SAM presents superior performance across all scenarios. These visualization results further validate the superiority of our M$^4$-SAM.

\subsection{Ablation Studies}
To validate the effectiveness of each component in our M$^4$-SAM, we conduct ablation studies on the RDVS dataset. %

\noindent{\bf Importance of Depth Information.}
Table~\ref{tab:ablation_depth_input} investigates the contribution of depth modality through two pseudo-depth variants.
When the RGB image is directly used as pseudo depth input (``Pseudo (Copy)''), the E-measure drops from 0.927 to 0.858, indicating that depth input provides complementary geometric information to enhance detection accuracy.
Using an all-zero depth map (``Pseudo (Black)'') slightly mitigates this drop, achieving E-measure of 0.895. %
Overall, these results confirm the necessity of depth information.

\begin{table}
  \centering
  \caption{Comparison of different parameter-efficient fine-tuning strategies.}
  \label{tab:ablation_peft}
  \resizebox{0.9\columnwidth}{!}{%
  \begin{tabular}{l|c|cccc}
    \hline
    Method & \begin{tabular}[c]{@{}c@{}}Mem.\\(GB)\end{tabular} &
    $E_\xi \uparrow$ & $S_\alpha \uparrow$ & $F_\beta \uparrow$ & $M \downarrow$ \\
    \hline
    None & 11.37 &
    0.839 & 0.816 & 0.693 & 0.033 \\ %

    Adapter & 14.66 &
    0.904 & 0.861 & 0.759  & 0.031 \\ %

    LoRA & 14.25 &
    0.904 & 0.864  & 0.764 & 0.031 \\ %
    Conv-LoRA & 11.60  &
    0.900 & 0.861 & 0.761 & 0.032 \\ %
    \rowcolor{red!20}
    \textbf{Ours} & 11.62 &
    \textbf{0.927} & \textbf{0.878} & \textbf{0.802} & \textbf{0.027} \\ %
    \hline
  \end{tabular}
  }
\vspace{-0.3cm}
\end{table}

\noindent{\bf Parameter-efficient fine-tuning strategies.}
Table~\ref{tab:ablation_peft} compares different parameter-efficient fine-tuning (PEFT) strategies for adapting SAM to RGB-D VSOD, where the ``Mem.'' column indicates training GPU memory usage for a single batch.

The frozen SAM encoder without any PEFT strategies (``None'') achieves the lowest performance across all metrics, highlighting the necessity of fine-tuning the encoder.
Although Adapter and LoRA improve performance, their lack of modality-specific design requires adopting a dual-encoder architecture. This leads to significant memory overhead during the training stage (14.66GB and 14.25GB).
Conv-LoRA mitigates the memory issues by incorporating a MoE structure, but its shallow experts design limits its capacity for effective multi-modal fusion, resulting in suboptimal performance (0.900 in E-measure).
Our method outperforms all alternatives, indicating the effectiveness of the proposed Modality-Aware MoE-LoRA.

\begin{table}
  \centering
  \setlength{\tabcolsep}{3.5 mm}
  \caption{Ablation studies of top-$K$ in MoE Gating.}
  \label{tab:ablation_topk}
  \resizebox{0.85\columnwidth}{!}{%
  \begin{tabular}{l|cccc}
    \hline
    top-$K$ & $E_\xi \uparrow$ & $S_\alpha \uparrow$ & $F_\beta \uparrow$ & $M \downarrow$ \\
    \hline
    1 & 0.903 & 0.857 & 0.766 & 0.031 \\
    \rowcolor{red!20}
    \textbf{2} & \textbf{0.927} & \textbf{0.878} & \textbf{0.802} & \textbf{0.027} \\
    3 & 0.914 & 0.865 & 0.775 & 0.030 \\
    \hline
  \end{tabular}
  }
\vspace{-0.3cm}
\end{table}

We also conduct ablations of the top-$K$ selection in MoE Gating. 
As shown in Table~\ref{tab:ablation_topk}, when reducing the $K$ to 1, the E-measure drops to 0.903, indicating the insufficient feature diversity from activating only one expert. 
Conversely, activating all experts causes feature redundancy, leading to a performance drop.
Overall, the best performance is achieved when $K$ is set to 2.
The results demonstrate that our Modality-Aware MoE-LoRA is well designed to balance specialization and complementarity among expert groups for effective multi-modal fusion.

\begin{table}
  \centering
  \setlength{\tabcolsep}{0.8 mm}
  \caption{Ablation studies of Pseudo-Guided Temporal Memory.}
  \label{tab:ablation_memory}
    \resizebox{1\columnwidth}{!}{%
  \begin{tabular}{l|cccc}
    \hline
    Models & $E_\xi \uparrow$ & $S_\alpha \uparrow$ & $F_\beta \uparrow$ & $M \downarrow$ \\
    \hline
    Baseline & 0.908 & 0.874 & 0.775 & 0.030 \\ %
    Baseline+Mem & 0.910 & 0.871 & 0.787 & 0.029 \\ %
    
    \rowcolor{red!20}
    \textbf{Baseline+Mem+Gated-MLF} & \textbf{0.927} & \textbf{0.878} & \textbf{0.802} & \textbf{0.027} \\
    \hline
  \end{tabular}
    }
\vspace{-0.3cm}
\end{table}

\begin{table}
  \centering
  \caption{Ablation studies of Decoder Feature Input for Memory.}
  \label{tab:ablation_gated_mlf}
    \resizebox{0.85\columnwidth}{!}{%
  \begin{tabular}{cc|cccc}
    \hline
    \multicolumn{2}{c|}{Decoder Feature} & \multirow{2}{*}{$E_\xi \uparrow$} & \multirow{2}{*}{$S_\alpha \uparrow$} & \multirow{2}{*}{$F_\beta \uparrow$} & \multirow{2}{*}{$M \downarrow$} \\ \cline{1-2}
    $X_D^1$ & $X_D^2$ \\
    \hline
    
    \Checkmark &            & 0.918 & 0.872 & 0.775 & 0.029 \\
    \rowcolor{red!20}
               & \Checkmark & \textbf{0.927} & \textbf{0.878} & \textbf{0.802} & \textbf{0.027} \\
    \Checkmark & \Checkmark & 0.909 & 0.871 & 0.787 & 0.029 \\
    \hline
  \end{tabular}
    }
\vspace{-0.3cm}
\end{table}

\noindent{\bf Pseudo-Guided Temporal Memory.}
To analyze the contribution of our temporal memory design, we conduct ablation studies as shown in Table~\ref{tab:ablation_memory}.
The baseline model, which directly uses coarse mask $P_c^1$ from the decoder as the prediction, achieves 0.908 in terms of E-measure.
Adding the Memory bank (``Baseline+Mem'') alleviates the performance drops, highlighting both the necessity of temporal information modeling and the effectiveness of our memory components.
When combined with Gated-MLF for multi-scale feature fusion, the E-measure significantly improves to 0.927, demonstrating that the utilization of multi-scale features is crucial for RGB-D VSOD.

Besides, it has been observed that the choice of feature levels fed into the memory module can significantly impact the segmentation performance~\cite{shao2024explore}. Therefore, we further investigate the optimal decoder feature level for Gated-MLF.
As shown in Table~\ref{tab:ablation_gated_mlf}, we can see that using $X_D^2$ alone achieves superior performance when compared to using $X_D^1$ alone or combining both.
A plausible explanation is that the middle-level feature $X_D^2$ preserves sufficient fine-grained details for accurate localization, providing a better balance between spatial details and semantic richness. 
This balance is essential for effective temporal modeling and multi-scale feature fusion in RGB-D VSOD task.

\begin{table}
  \centering
  \setlength{\tabcolsep}{3.5 mm}
  \caption{Ablation studies of training clip length $T$ .}
  \label{tab:ablation_vidlen}
  \resizebox{0.75\columnwidth}{!}{%
  \begin{tabular}{l|cccc}
    \hline
    $T$ & $E_\xi \uparrow$ & $S_\alpha \uparrow$ & $F_\beta \uparrow$ & $M \downarrow$ \\
    \hline
    2 & 0.897 & 0.854 & 0.750 & 0.033 \\
    \rowcolor{red!20}
    \textbf{4} & \textbf{0.927} & \textbf{0.878} & \textbf{0.802} & \textbf{0.027} \\
    6 & 0.908 & 0.875 & 0.778 & 0.031 \\
    \hline
  \end{tabular}
  }
\vspace{-0.5cm}
\end{table}

\noindent{\bf Number of Frames per Clip.}
Table~\ref{tab:ablation_vidlen} examines the impact of training clip length $T$ on temporal modeling capability.
When reducing $T$ to 2, the variant results in limited performance (0.897 in E-measure), indicating that such a short clip length is insufficient to capture temporal dependencies.
Conversely, extending $T$ to 6 also results in degraded performance.
The $T{=}4$ configuration strikes the best balance between temporal modeling granularity and training efficiency, and is adopted as the default for both training and inference, where varying the memory size from $T{=}4$ to $T{=}32$ at test time yields stable accuracy.

\section{Conclusion}
\label{sec:conclusion}

In this paper, we present a prompt-free framework M$^4$-SAM, which adapts SAM2 for RGB-D video salient object detection.
To address the challenges of conventional PEFT methods, including the lack of spatial priors, insufficient employment of SAM's multi-scale features, and reliance on manual prompts for initialization, we deploy three key designs: 
the Modality-Aware MoE-LoRA, which enhances LoRA with convolutional experts and modality-specific routing, achieving adaptive RGB-D feature fusion and efficient fine-tuning; 
the Gated Multi-Level Feature Fusion, which hierarchically combines multi-scale encoder features guided by the decoder's mid-level feature, balancing fine-grained details with semantic information;
and the Pseudo-Guided Initialization, which bootstraps temporal memory using a pseudo mask, eliminating the reliance on manually provided prompts.
Extensive experiments on three RGB-D VSOD benchmarks demonstrate that M$^4$-SAM achieves state-of-the-art performance, offering a new solution to efficiently adapt SAM2 for the RGB-D VSOD task.
In future work, we will further optimize our model's generalizability and extend our approach to more downstream multi-modal video understanding tasks.

\section*{Acknowledgements}
This work was supported by the National Natural Science Foundation of China (No.62271180, 62471278, 62471285).

{
    \small
    \bibliographystyle{ieeenat_fullname}
    \bibliography{main}
}

\end{document}